\documentclass{article}

\usepackage{arxiv}

\usepackage[utf8]{inputenc} 
\usepackage[T1]{fontenc}    
\usepackage{hyperref}       
\usepackage{url}            
\usepackage{booktabs}       
\usepackage{amsfonts}       
\usepackage{nicefrac}       
\usepackage{microtype}      
\usepackage{lipsum}
\usepackage{graphicx}
\usepackage{subfigure}
\usepackage{bm}
\usepackage{amssymb}
\usepackage{amsthm}
\usepackage{latexsym}
\usepackage{bbding}
\usepackage{algorithm}
\usepackage{algorithmic}
\usepackage{booktabs}
\usepackage{multirow}
\usepackage{url}
\usepackage{array}
\usepackage{amsmath}
\usepackage{booktabs}
\usepackage{xcolor}
\usepackage{epstopdf}
\usepackage{tabularx}

\title{Privacy-Preserving Federated Learning on Partitioned Attributes}

\author{
	Shuang Zhang, Liyao Xiang, Xi Yu, Pengzhi Chu, Yingqi Chen\\
	Shanghai Jiao Tong University \\
	\And
	Chen Cen, Li Wang \\
	Ant Group \\
}
\begin{document}
\maketitle

\begin{abstract}
Real-world data is usually segmented by attributes and distributed across different parties. Federated learning empowers collaborative training without exposing local data or models. As we demonstrate through designed attacks, even with a small proportion of corrupted data, an adversary can accurately infer the input attributes. We introduce an adversarial learning based procedure which tunes a local model to release privacy-preserving intermediate representations. To alleviate the accuracy decline, we propose a defense method based on the forward-backward splitting algorithm, which respectively deals with the accuracy loss and privacy loss in the forward and backward gradient descent steps, achieving the two objectives simultaneously. Extensive experiments on a variety of datasets have shown that our defense significantly mitigates privacy leakage with negligible impact on the federated learning task.
\end{abstract}


\section{Introduction}
Many machine learning applications and systems today are driven by huge amounts of data distributed across different entities. A large proportion of the data is proprietary to the owner and is prohibited from sharing. For the distributed data, it has become a practice to perform federated learning \cite{konevcny2016federated1,konevcny2016federated2,mcmahan2017communication,yang2019federated} which builds a common model without each entity exposing their private data. For example, customer behavior prediction would require a combination of purchase history from the e-commerce platform, credit score from banks, paychecks from employees, etc. Different from centralized learning, attributes belonging to the same data item is partitioned across different entities in vertical federated learning. Each entity processes its local data by the local model, and sends the intermediate outputs (DNN features) to a central server to build a joint model. 

The intermediate outputs exchanged pose potential threats to the privacy of each participating entity. To preserve data privacy, a series of homomorphic encryption-based methods have been proposed such as \cite{cheng2019secureboost,wu2020privacy,zhang2018gelu}. However, homomorphic encryption does not well support non-linear functions which are common in DNNs, letting alone being highly expensive in computation. Hence much of the intermediate output exchanged is in plaintext \cite{cheng2019secureboost,liu2020federated}, leaving its privacy threats an open problem. From the machine learning perspective, recent studies have shown that for a well-trained model, the model parameters or the DNN features contain much more input information than one would expect \cite{dosovitskiy2016inverting,mahendran2015understanding,fredrikson2015model,melis2019exploiting,nasr2019comprehensive}. However, it is not well understood when the model has not converged, how much private information is revealed by the intermediate outputs.

To demonstrate the feature leakage, we introduce privacy attacks to the partitioned attributes in federated learning. We assume the adversary is able to corrupt a small proportion of the inputs, and obtain the corresponding intermediate outputs, based on which it trains a {\em decoder} reconstructing other private inputs. Attacks differ by means that the adversary trains its decoder: it is straightforward to train a decoder given features generated in the same iteration of update, but when insufficient amounts of features are collected, the decoder needs to transfer across features generated in successive iterations. We illustrate that, even with a very small percentage of corrupted data, the attack success rate is exceedingly high. Later we show the attack performance can also serve as the evaluation metric for the privacy-preserving capability of DNNs.

We propose countermeasures to mitigate the privacy leakage. A naive approach is to include a privacy loss as a training objective. However, reaching a state where the privacy loss is minimized does not mean privacy is guaranteed in the training process. The main difficulty lies in that the intermediate output is exposed in each iteration and thus requiring privacy restrictions on each release of them. Hence we design an adversarial learning approach for releasing the intermediate outputs. We have the local model pit against the decoder by tuning its weights against the reconstruction loss. But the defense unfortunately leads to accuracy performance decline for the federated learning task --- the seemingly contradictory objectives of accuracy and privacy are difficult to balance.

To guarantee accuracy and privacy at the same time, we propose our privacy-preserving federated learning algorithm on the basis of forward-backward splitting (FBS) \cite{goldstein2014field}. We split up the optimization loop for accuracy loss and privacy loss: the forward gradient descent step seeks a direction to reduce the accuracy loss and obtains a model ${\theta}$; the backward gradient descent searches a direction which minimizes the privacy loss in the vicinity of ${\theta}$. It resolves two problems: first, the sub(gradient) of the privacy loss is intrinsically hard to express as it evolves with the decoder; second, the convergence of FBS is important to simultaneously achieve the accuracy and privacy objectives. Moreover, for better balancing the two goals, we multiply weight factors with the objectives and automatically adjust them in training.

Highlights of our contributions are as follows. {\em First,} by our privacy attacks, we demonstrate federated learning over partitioned attributes are vulnerable such that inputs can be inferred from the intermediate outputs. {\em Second,} we propose a defense approach by adversarially learning an intermediate representation against the worst-case decoder. To prevent accuracy decline, we further enhance the defense in the framework of forward-backward splitting, which optimizes the accuracy and privacy objective in an alternate fashion. {\em Last,} by extensive experiments in federated learning, we show that our privacy attack reveals a significant amount of private information even with a very small proportion of corrupted data; with our proposed defense, the privacy leakage is mitigated to around halfway between that without any defense and the random guesses. Our methods demonstrate promising applicability in real-world settings.

\section{Related Work}

Most previous works have demonstrated the threat of privacy leakage in horizontal federated learning. \cite{nasr2019comprehensive,melis2019exploiting} launch the membership inference attack which aims to determine whether a specific record is in a client’s training dataset or not. Other attacks including \cite{melis2019exploiting,geiping2020inverting,zhu2019deep,bagdasaryan2020backdoor} exploit the iteratively exchanged gradients during training. \cite{melis2019exploiting} shows the periodically exchanged model updates leak unintended information about participants’ training data. Likewise, \cite{geiping2020inverting} reconstructs input images from gradients. \cite{zhu2019deep} introduces an approach to obtain the local training data from public shared gradients. Different from them, \cite{luo2020feature} proposes feature inference attack to vertical FL but at the inference stage, rather than at the training phase.

Mainstream methods for privacy-preserving federated learning include differential privacy and crypto-based solutions. Differential privacy introduces additive noise to the features or the model \cite{geyer2017differentially,roy2020crypt} for privacy but degrades model performance. Homomorphic encryption (HE) supports model updates exchange in the encrypted domain without exposing any data in plaintext. \cite{hardy2017private} uses Taylor expansion to approximate the Sigmoid function and \cite{zhang2018gelu} splits each neuron into linear and nonlinear components and implement them separately on non-colluding parties. However, it is not feasible to approximate a deep neural network with a low-degree polynomial function accurately and efficiently, letting alone in the encrypted domain of high complexity.

\section{Denotations and Background}

\textbf{Vertical Federated Learning (VFL).}\label{sec:vfl}
Consider a set of clients $\mathcal{M}:\{1,\cdots,M\}$ collaboratively training a model on $N$ data samples $\{\mathbf{x}_i,y_i\}_{i=1}^N$. The $i$-th data item $\mathbf{x}_i \in \mathbb{R}^{d}$ is distributed across $M$ clients $\{\mathbf{x}_i^{m} \in \mathbb{R}^{d_m}\}_{m=1}^M$, where $d_m$ is the number of raw data attributes hosted by client $m$. Before the training procedure, all clients would align their respective attributes for the same data item. The clients share one common top model $\theta_T$ which is hosted either by an arbiter, or one of the clients, along with the labels $\mathbf{y}:=\{y_i\}_{i=1}^N$. Each client owns a local model $\theta_m$ as well as the local dataset $\mathcal{D}_m:=\{\mathbf{x}_i^m\}_{i=1}^N$ for $m \in \{1, \cdots, M\}$. The dataset is jointly represented as $\mathcal{D}:=\{\mathcal{D}_m\}_{m=1}^M$.

\begin{figure*}[htbp]
	\centering
	\includegraphics[width=0.95\linewidth]{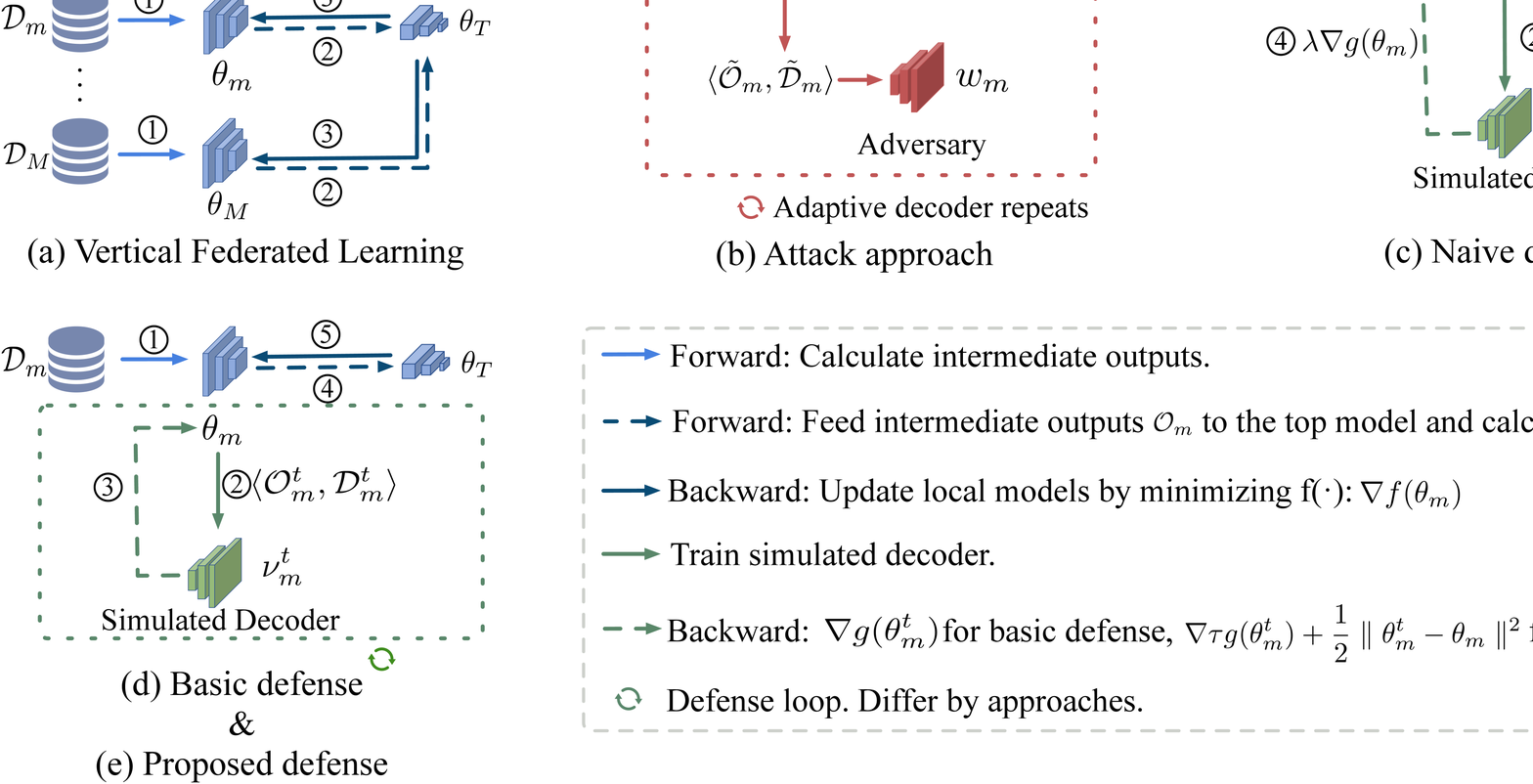}
	\centering
	\caption{The framework of vertical federated learning.}
	\label{fig:vertical_framework}
\end{figure*}

Letting $f(\cdot)$ be the loss of the predicted outputs against the ground truth labels, the objective of VFL is
\begin{equation} \label{eq:vfl}
\mathop{\text{minimize}}_{\theta_1,\cdots,\theta_M,\theta_T} f(\theta_1,\cdots,\theta_M,\theta_T;\mathcal{D}, \mathbf{y}).
\end{equation}
The training procedure of VFL is shown in Fig.~\ref{fig:vertical_framework}(a). At each iteration, client $m$ feeds its own data $\mathcal{D}_m$ to its model to produce the intermediate output $\mathcal{O}_m$. It sends $\mathcal{O}_m$ to the top model for further forward computation. At the end of forward loop, the top model calculates the accuracy loss on the collective features from all clients. Then the top model updates $\theta_{T}$ and propagates the error backward to each client, with which each client updates its own local model $\theta_{m}$. The new iteration repeats the above steps until convergence.

\textbf{Forward-Backward Splitting (FBS).} 
FBS provides a practical solution for non-differentiable and constrained convex optimization in the following form:
\begin{equation}\label{eq:fbs}
\text{minimize} ~f(x) + g(x)
\end{equation}
where $f$ is convex and differentiable and $g$ can be neither differentiable nor finite-valued. Hence Eq.~\ref{eq:fbs} cannot be solved by simple gradient-descent methods. In each iteration, FBS performs a simple forward gradient descent step on $f$ by computing
$$\hat{x}^{k+1} = x^k - \tau^k\nabla f(x^k),$$
and then calculates the proximal operator:
\begin{equation}
x^{k+1} = \text{prox}_g(\hat{x}^{k+1},\tau^k) = \arg \min \limits_{x} \tau^k g(x) + \frac{1}{2} \parallel x - \hat{x}^{k+1} \parallel^2
\end{equation}
as the backward descent step on $g$ where $\tau^k$ is the step size. It is called `backward' because the sub-gradient of $g$ is evaluated at the final point $x^{k+1}$ rather than the starting point.

FBS is proved to converge when the problem is convex and the step size $\tau^k$ satisfies certain stability bounds. Although lacking theoretical guarantees for the non-convex problems, FBS still works quite well in practice \cite{goldstein2014field}.
\section{Proposed Approaches}
We first introduce the threat model in the VFL, and reveal the potential privacy leakage that the client faces. As the main contribution, we show our proposed countermeasures against such privacy leakage.

\textbf{Threat model.}
We assume that the top model is controlled by an honest-but-curious adversary who is interested in peeking clients' private inputs. It has access to the intermediate outputs of the victim and is able to acquire a fairly small percentage of the victim's inputs. The assumption is realistic as the adversary may corrupt the data source of the victim by injecting fake data entries, or secretly intercept the victim's private inputs. For example, if the victim participant collects consumers' purchasing history, the adversary can register as a consumer and perform artificial transactions with the victim participant. We refer to those corrupted data as {\it poison data}, and they are retrieved at the item alignment step before training. By gathering the poison data and the corresponding intermediate outputs, the adversary constructs a training dataset for its {\it decoder}. The adversary trains the decoder to recover other private data of the victim. The question we ask is that, given the fraction of poison data, how much private information could the adversary infer about the victim?

\subsection{Attacks}\label{sec:attack}
To answer the question, we assume the adversary collects victim $m$'s poison data as $\mathcal{\tilde{D}}_m$ and the corresponding intermediate outputs as $\mathcal{\tilde{O}}_m$. The model parameter of the decoder is represented as $w_{m}$. Letting the reconstruction loss be $l(\cdot)$, the objective of the decoder is:
\begin{equation}\label{eq:attack}
\mathop{\text{minimize}}_{w_{m}} l(w_{m}; \mathcal{\tilde{O}}_m, \mathcal{\tilde{D}}_m).
\end{equation}
Essentially, the adversary aims to train $w_{m}$ over the feature-data pairs $\langle \mathcal{\tilde{O}}_m, \mathcal{\tilde{D}}_m \rangle$ and obtains a decoder which returns a reconstructed input when fed with intermediate features. Then the decoder is used to recover other sensitive features. 

We propose two types of attacks, respectively a {\bf static attack} and an {\bf adaptive attack} shown in Fig.~\ref{fig:vertical_framework}(b), depending on how the adversary trains its decoder. In a static attack, the adversary collects poison data all at one iteration, and trains a decoder particularly for this iteration. The attack suits the case where sufficient poison data is gathered in a single iteration. More often, the poison data is scattered in different training batches and the adversary has to collect them over multiple iterations. In that case, the adversary can launch an adaptive attack, training the decoder over stream of poison data and features. Since the adaptive attack trains the decoder across different states of the local model, it relies on the transferability of the decoder. We will compare the strength of the two attacks in the experiments.

{\bf Privacy leakage measure.} 
The measurement we take for privacy leakage is the reconstruction loss of the adversary. It can be the mean squared difference between the reconstructed input and the original one, or the error rate of inferring a particular attribute. The privacy metrics differ according to the specific task of the adversary, which can be a regression task or a classification one, depending on the data types.

\subsection{Defenses}
To fight against privacy breaching, we propose a client-side defense approach. Ideally, the intermediate outputs of the client should reveal nothing about the inputs, but contain sufficient information to make the correct prediction. Hence a naive approach is to take the privacy objective into consideration along with the accuracy loss in training. In our defense, we assume a simulated worst-case adversary who has full access to the client's local model, all training data and the corresponding intermediate outputs. Similar to the real-world adversary described in Sec.~\ref{sec:attack}, the simulated adversary trains a decoder $\nu_{m}$ over the training set to recover the inputs to minimize $l(\cdot)$. Note that the simulated adversary is the worst since it has the full dataset rather than a few poison samples. The client's objective is to minimize the accuracy loss while maximizing the reconstruction loss against the worst-case decoder. For consistency, we denote the privacy loss as $g(\cdot)$ which is a non-increasing function of $l(\cdot)$. Formally, the overall objective is
\begin{equation}\label{eq:defense}
\mathop{\text{minimize}}_{\theta_1,\cdots,\theta_M,\theta_T} f(\theta_1,\cdots,\theta_M,\theta_T) + \\ \lambda \sum_{m=1}^{M} g(\theta_m, \nu_{m}^*),
\end{equation}
where
\begin{equation}\label{eq:fake_decoder}
\nu_{m}^* = \arg \min \limits_{\nu_{m}} l(\nu_{m}; \mathcal{O}_m, \mathcal{D}_m).
\end{equation}

In our {\bf naive defense} (Fig.~\ref{fig:vertical_framework}(c)), client $m$ feeds its own data $\mathcal{D}_m$ to its local model to produce the intermediate outputs $\mathcal{O}_{m}$. Given the feature-data pairs, the decoder obtains the optimized parameters by Eq.~\ref{eq:fake_decoder}. In the backward loop, the client computes gradients $\nabla f(\theta_m) + \lambda \nabla g(\theta_m)$ (of Eq.~\ref{eq:defense}) to update its local model. While being intuitive, the naive defense is not effective. As the gradients associated with the accuracy and privacy losses are computed in the backward loop, the defense for the intermediate outputs always lags one iteration behind. That is, the intermediate outputs are uploaded without any posterior privacy concern. 

In fact, we can separate the two objectives: the model update against the accuracy loss remains in the forward-backward loop while inserting the local model update against the privacy loss in between the accuracy loops. In particular, we need to guarantee privacy for the intermediate outputs before uploading them. Hence we propose a defense approach detaching the privacy goal from the original objective.

Our {\bf basic defense} is given in Fig.~\ref{fig:vertical_framework}(d). In each training iteration, the client first optimizes its local model against the privacy objective: it trains the simulated decoder to obtain $\nu_{m}^{*}$ by Eq.~\ref{eq:fake_decoder} and the client $m$ updates its local model by minimizing the privacy loss:
\begin{equation}\label{eq:basic_defense}
\mathop{\text{minimize}}_{\theta_{m}} g(\theta_{m},\nu_{m}^{*}; \mathcal{O}_m, \mathcal{D}_m).
\end{equation}
Assuming the local model obtained in this step is $\tilde{\theta}_{m}$, the client uploads the intermediate output produced on $\tilde{\theta}_{m}$ to the top model. The rest of the forward and backward computation is the same as in the typical VFL (Eq.~\ref{eq:vfl}). Within each iteration, the client updates its local model twice: the first is for its privacy goal and the second is for accuracy objective.


However, such a basic defense may not work in practice. The client only has a local view on its data and the local model, but no idea about others'. Hence it cannot control the accuracy loss while optimizing its privacy objective. It is very likely that the obtained $\tilde{\theta}_{m}$ leads to accuracy drop. Therefore, the accuracy of the VFL task would be severely affected, or even the model does not converge. However, it is not possible to train for the accuracy loss and privacy loss at the same time, since the client does not get the accuracy feedback from the top model in the privacy training loop.
\begin{algorithm}
	\renewcommand{\algorithmicrequire}{\textbf{Input:}}
	\renewcommand{\algorithmicensure}{\textbf{Output: }}
	\caption{Minimax}
	\begin{algorithmic}[1]
		\REQUIRE Client $m$'s current parameter $\theta_{m}$.
		\ENSURE  Client $m$'s new parameter $\tilde{\theta}_{m}$.
		\STATE Initialize $\hat{\theta}_{m}^1 = \theta_{m}$
		\FOR {$t = 1$ to $N_2$}
		\STATE Update the simulated decoder by ascending stochastic gradient for $M_1$ steps:
		
		$\nu_{m}^{t+1} \leftarrow \nabla g(\hat{\theta}_m^t, \nu_{m}^{t};\mathcal{O}_m, \mathcal{D}_m)$
		\STATE Update the client $m$ by descending stochastic gradient for $m_2$ steps.
		
		$\hat{\theta}_{m}^{t+1} \leftarrow \nabla_{\hat{\theta}_{m}^t} \tau^t g(\hat{\theta}_m^t, \nu_{m}^{t+1};\mathcal{O}_m, \mathcal{D}_m) + \frac{\parallel \hat{\theta}_{m}^t - \theta_{m}\parallel^2}{2}$
		
		$\tau^{t+1}\leftarrow \nabla_{\tau^t} \tau^t g(\hat{\theta}_m^t, \nu_{m}^{t+1};\mathcal{O}_m, \mathcal{D}_m) + \frac{\parallel \hat{\theta}_{m}^t - \theta_{m}\parallel^2}{2}$
		
		\ENDFOR
		\RETURN $\tilde{\theta}_{m} = \hat{\theta}_{m}^{N_2}$
	\end{algorithmic}
	\label{alg:minimax}
\end{algorithm}

To resolve the issue, we propose a defense as in Fig.~\ref{fig:vertical_framework}(e). We observe that the forward-backward splitting (FBS) \cite{goldstein2014field} algorithm naturally splits the optimization procedure for two objectives --- the forward gradient descent step only deals with the first objective whereas the backward gradient descent step only involves the other. The FBS has a nice convergence property when the stepsize satisfies certain stability bounds. Therefore, in Eq.~\ref{eq:defense}, we deal with the accuracy loss $f(\cdot)$ in the forward gradient descent step and the privacy loss $g(\cdot)$ in the backward gradient descent step. In the backward step, we compute the proximal operator as follows:
\begin{equation}\label{eq:proximal}
\text{prox}_g(\theta,\tau) = \arg \min \limits_{\theta_{m}} \tau g(\theta_m, \nu_{m}^*;\mathcal{O}_m, \mathcal{D}_m) + \frac{1}{2} \parallel \theta_{m} - \theta \parallel^2
\end{equation}
where $\tau$ is the stepsize and $\nu_{m}^*$ is achieved by solving Eq.~\ref{eq:fake_decoder}. The proximal operator finds a point close to the minimizer of $g$ without straying too far from the starting point $\theta$. Since the proximal operator is obtained w.r.t. $\nu_{m}^*$,  and $\nu_{m}^*$ can be obtained by maximizing $g(\cdot)$, $\nu_{m}^*$ in fact does not represent an optimal solution but a saddle point. We illustrate the minimax procedure in Alg.~\ref{alg:minimax} where the model weights $\theta_{m}$ and $\nu_{m}$ are optimized alternatively. $N_2$ is the defense epoch number. Moreover, since the efficiency of forward-backward splitting algorithm is sensitive to $\tau$, we treat $\tau$ as a hyper-parameter and update it automatically. Alg.~\ref{alg:minimax} takes as inputs the client's local model weights updated by the accuracy loss and returns the weights updated by the proximal operator.
In practice, we adopt two hyper-parameters in Alg.~\ref{alg:minimax}: $\tau_1$ for $g(\cdot)$ and $\tau_2$ for $\frac{1}{2}\parallel \hat{\theta}_{m}^t - \theta_{m}\parallel^2$ and update them automatically. Apart from the $\ell_2$ norm of parameters, we can also use feature loss $\frac{1}{2}\parallel h(\hat{\theta}_{m}^t;\mathcal{D}_m) - h(\theta_{m};\mathcal{D}_m)\parallel^2$ as replacement where $h(\cdot)$ is the function mapping raw data to intermediate outputs.
\begin{algorithm}
	\renewcommand{\algorithmicrequire}{\textbf{Input:}}
	\renewcommand{\algorithmicensure}{\textbf{Output: }}
	\caption{Privacy-Preserving VFL}
	\begin{algorithmic}[1]
		\REQUIRE Models of $M$ clients $\theta_{1}, \ldots, \theta_{M}$; the top model $\theta_{T}$
		\FOR {$k = 1$ to $N_1$}
		\FOR {each client $m$}
		\STATE $\tilde{\theta}_{m}^k$ = Minimax($\theta_{m}^k$)
		\ENDFOR
		\STATE Update top model:
		
		$\theta_T^{k+1} = \theta_T^k- \eta\nabla f(\tilde{\theta}_{1},\cdots,\tilde{\theta}_{M},\theta_T;\mathcal{D})$
		\FOR {each client $m$}
		\STATE $\theta_{m}^{k+1} = \tilde{\theta}_{m}^k - \eta\nabla  f(\tilde{\theta}_{1},\cdots,\tilde{\theta}_{M},\theta_T;\mathcal{D})$
		\ENDFOR
		\ENDFOR
	\end{algorithmic}
	\label{alg:algorithm1}
\end{algorithm}

The overall algorithm is presented in Alg.~\ref{alg:algorithm1}. Line 5-8 is the forward step which performs gradient descent to minimize the accuracy loss. Line 2-4 is the backward step which minimizes the privacy loss against a worst-case decoder. $N_1$ is the epoch number of the overall VFL, and $\eta$ is the learning rate. Although there is no theoretical guarantee that the FBS would converge for non-convex loss functions, FBS would be no worse than others since no known method can guarantee optimality in polynomial time \cite{goldstein2014field}.

\section{Evaluation}
\begin{table*}[!htbp]
	\centering
	\scriptsize
	\caption{Datasets and Experimental Settings}
	\scalebox{1.15}{
		\begin{tabular}{ c | c | c | c | c }
			\toprule
			Dataset & Attribute Type & Attributes Number (Total/Attack/Defense) & Instances Number & Attack Task \\ \hline
			Purchase100 \cite{shokri2017membership} & Binary & 600/300/300 & 197324 & Classification / Regression \\ \hline
			Credit \cite{shokri2017membership} & Numerical & 28/1/1 & 284807 & Regression \\ \hline
			UCI Adult & Categorical & 10/1/1 & 48842 & Classification\\ \hline
			COCO-QA \cite{ren2015exploring} & Image,Text & $\sim$/90/90 & 78736 & Classification \\ 
			\bottomrule
	\end{tabular}}
	\label{tab:dataset}
\end{table*}

Our algorithm is implemented on {\tt PyTorch 1.5.0} and all experiments are done on Intel Xeon Processor with GPU GeForce RTX 2080 Ti.
\subsection{Settings}
\textbf{Datasets.} We pick four representative datasets (shown in Tab.~\ref{tab:dataset}) covering different types of attributes to validate our approaches. The default VFL task is classification involving two clients, one being the attacker who also controls the top model, and the other being victim. Purchase100, Credit and UCI Adult are tabular datasets whereas COCO-QA contains two attributes: images and texts. All attributes are split evenly across the clients.


\textbf{Privacy metrics.} We set different privacy-preserving goals for these datasets. For tabular data, the attack is to reconstruct one or several attributes. Depending on the data type, the attack task can be categorized as classification or regression. The attack to COCO-QA is to predict whether the image contains a certain object and thus it is a classification task. We consider only one attribute in Credit or UCI Adult as private, but all attributes in Purchase or COCO-QA need protection.

We evaluate privacy by the attack success rate of an attacker who possesses a given proportion of poison data, on the rest of the (unseen) train data. The default poison data ratio is $\alpha = 5 \%.$ The measure of the attack success rate vary per datasets. For categorical attributes, we use {\it error rate}; for datasets of which the attribute distribution is highly uneven (sparse), we adopt {\it recall/f1-score}; for numerical attributes, we use the {\it mean squared error (MSE)} between the reconstructed data and the ground truth as the privacy metric. Note that a high privacy level means unsuccessful attack, indicated by high error rate, low recall, high MSE, etc. Since training is iteratively performed on the same data, we record the {\em minimum privacy} level as the privacy leakage of the entire training process.

\textbf{Models.} Since tabular data is easy to train, we use combinations of fully-connected layers and activation layers as models for the local network, top model and the decoder. For COCO-QA, the local neural networks are VGG11 for images and LSTM for texts, and the top model only contains fully-connected layers. Its decoder contains de-convolution layers followed by convolution layers. By default, Adam optimizer is used for training with learning rate $0.0001$.

\subsection{Attack Results}
We show the performance of static and adaptive attacks and their impact factors. The privacy attack is overall successful. For example, with less than $1\%$ poison data, the attacker can achieve over $0.9$ recall on Purchase. Even on complex dataset COCO-QA (90 objects, highly sparse), with merely $5\%$ poison data, the recall is nearly $0.5$.

\begin{figure*}[!htbp]
	\centering  
	\subfigure[Attack and accuracy performance at each epoch of VFL on Purchase.]{
		\label{fig:attack_1}
		\includegraphics[width=0.20\textwidth]{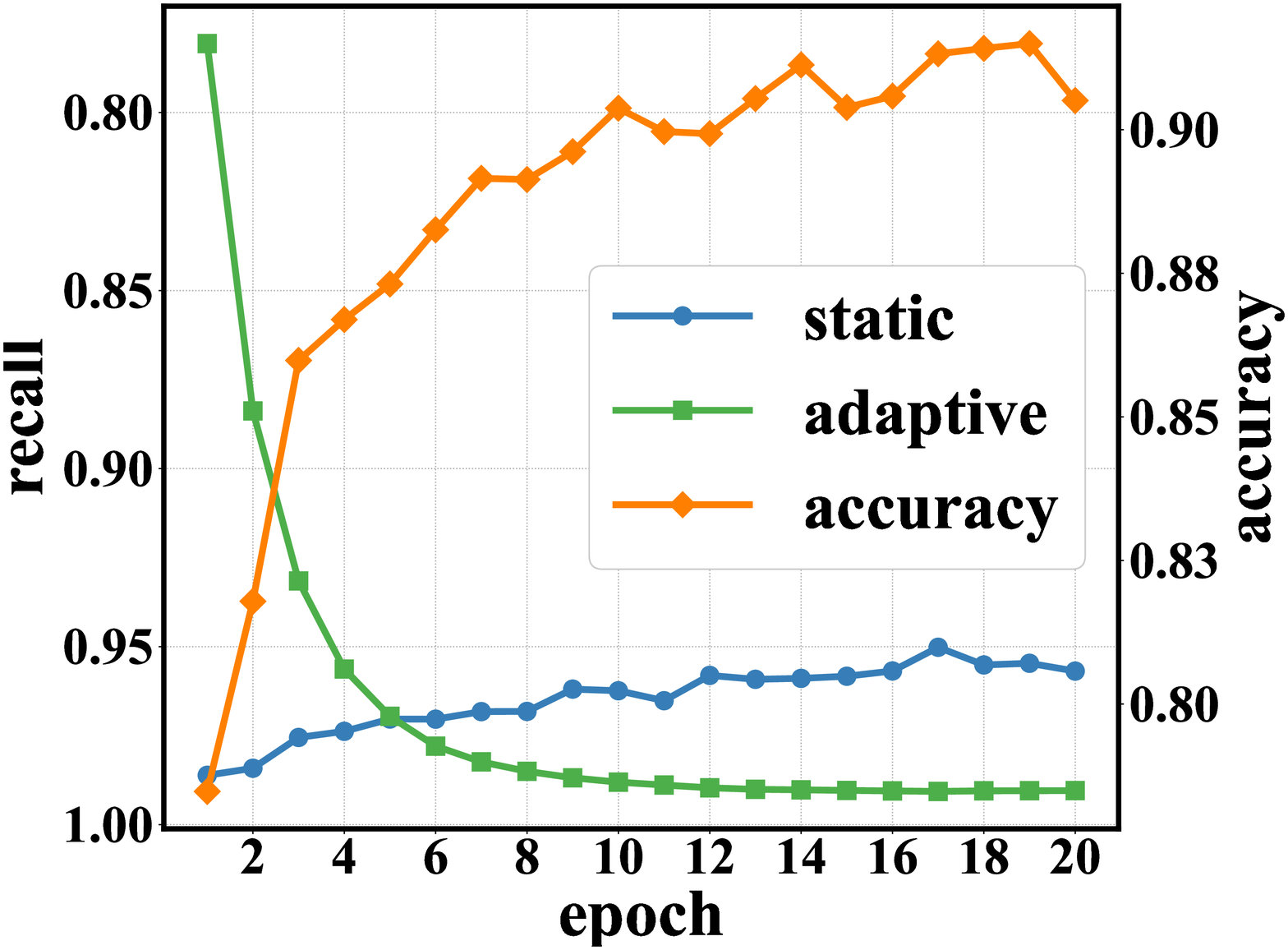}}
	\subfigure[Attack and accuracy performance at each epoch of VFL on COCO-QA.]{
		\label{fig:attack_2}
		\includegraphics[width=0.20\textwidth]{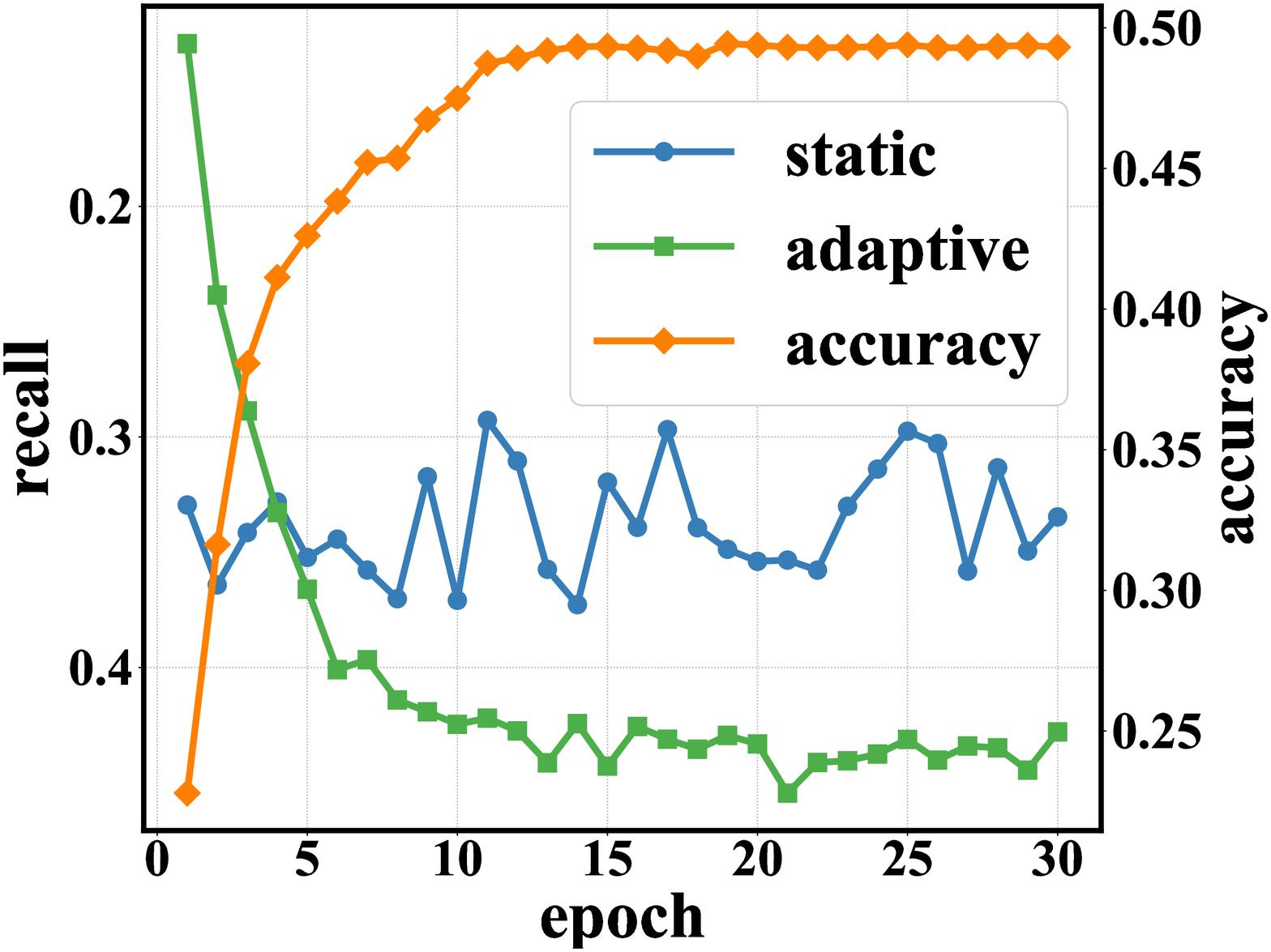}}
	\subfigure[Attack performance vs. poison ratio on Purchase.]{
		\label{fig:attack_3}
		\includegraphics[width=0.21\textwidth]{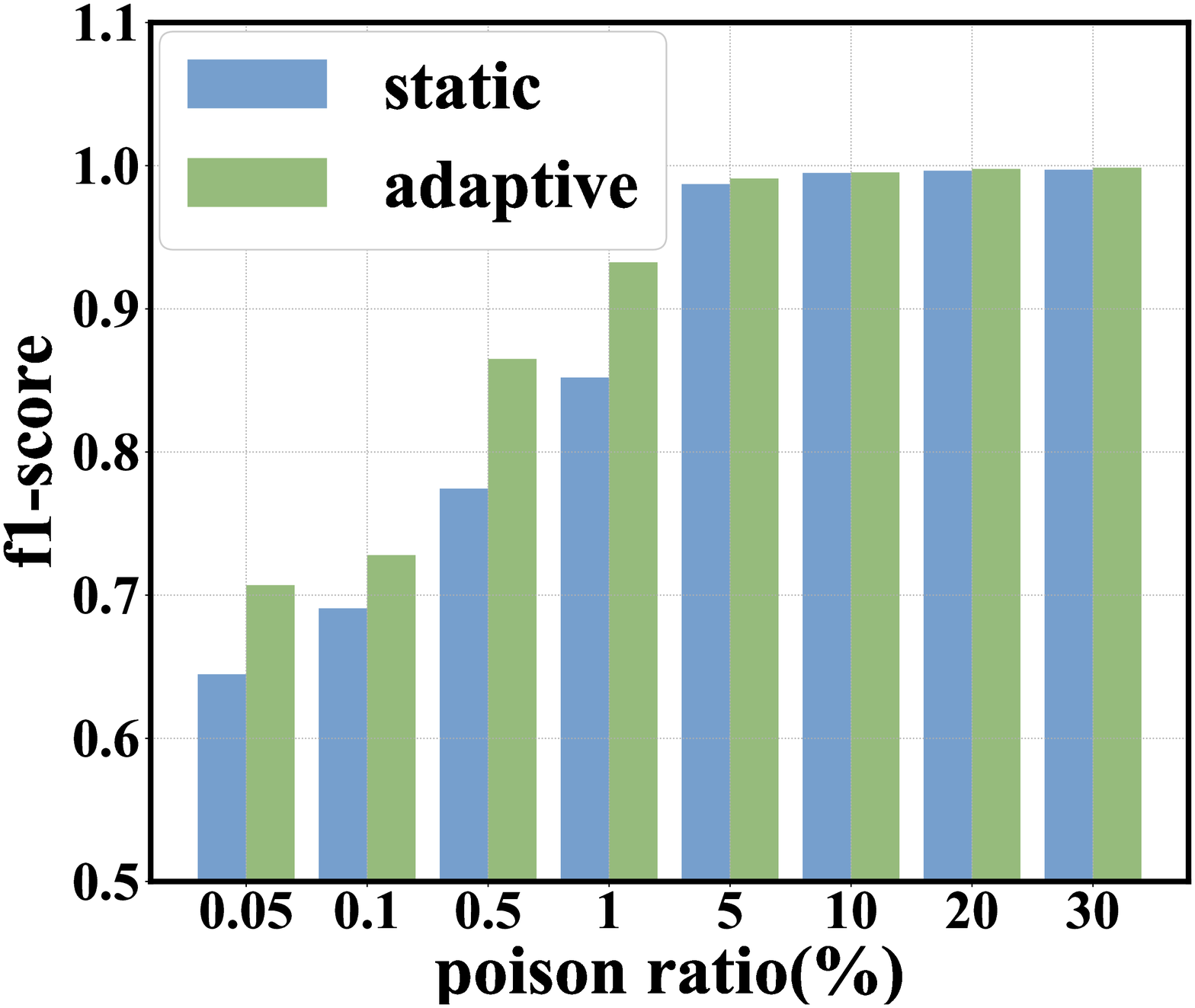}}
	\subfigure[Attack performance vs. poison ratio on COCO-QA.]{
		\label{fig:attack_4}
		\includegraphics[width=0.21\textwidth]{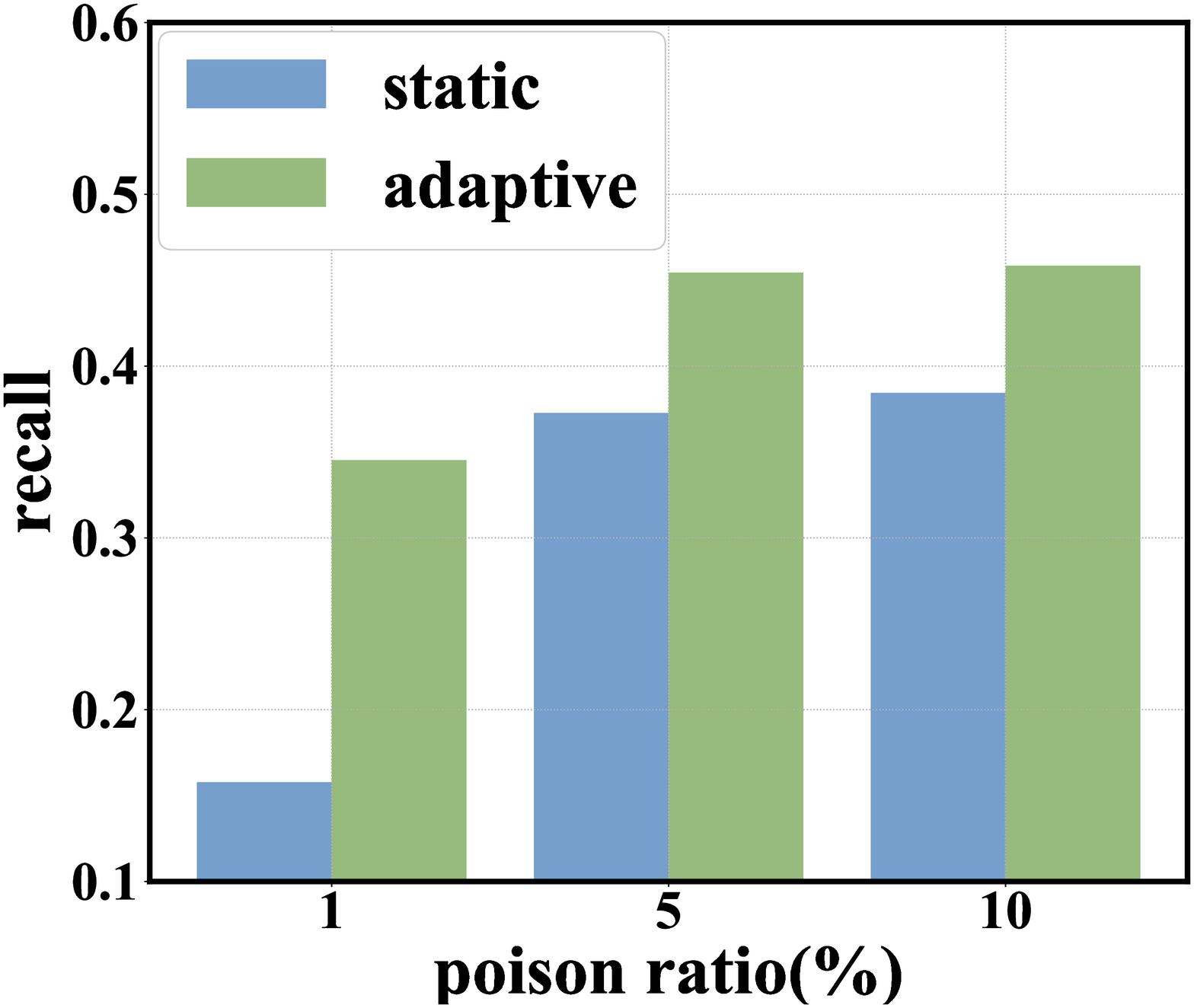}}
	\caption{Attack results on Purchase and COCO-QA.}
	\label{fig:purchase_all_attack}
\end{figure*}

Fig.~\ref{fig:attack_1} and Fig.~\ref{fig:attack_2} show how accuracy and privacy evolve throughout training. Each point on the static attack represents an independent attack launched at that epoch. We observe that, in most cases, the adaptive attack performs better than the static, mostly because more data is collected and the attack can transfer across features obtained in different rounds. On Purchase (Fig.~\ref{fig:attack_1}), the static attack performance gets slightly worse, indicating the features learned to be less revealing towards the end of training. An exception is Adult (Tab.~\ref{tab:batch_size}), where the static attack performs better than the adaptive, and it may be because the training batches vary significantly resulting in lack of transferability.
\begin{table*}[!htbp]
	\centering
	\scalebox{1.00}{\begin{tabular}{ c | c | c | c | c }
			\toprule
			Dataset & Batch size & 32 & 128 & 256 \\ \hline
			\multirow{2}{*}{\begin{tabular}[c]{@{}c@{}}Purchase\\ (Recall)\end{tabular}}  & Static     & 0.9861 & 0.9883 & 0.9918 \\ \cline{2-5} 
			& Adaptive & 0.9907 & 0.9973 & 0.9984 \\ \hline
			\multirow{2}{*}{\begin{tabular}[c]{@{}c@{}}Adult\\ (Error rate)\end{tabular}} & Static     & 0.0130 & 0.0131 & 0.0110 \\ \cline{2-5} 
			& Adaptive & 0.0193 & 0.0185 & 0.0160 \\
			\bottomrule
	\end{tabular}}
	\caption{Attack performance vs. batch size on Purchase and Adult.}
	\label{tab:batch_size}
\end{table*}

{\bf Poison ratio.}
Fig.~\ref{fig:attack_3} and Fig.~\ref{fig:attack_4} show that attack performance enhances as $\alpha$ increases, which is obvious as more training data is collected. When $\alpha$ is small, adaptive attack is stronger than the static one but the gap diminishes with the increase of $\alpha$, meaning that the static decoder can collect sufficient data for training when $\alpha$ surpasses a threshold.

{\bf Batch size.}
Tab.~\ref{tab:batch_size} demonstrates that as the batch size increases, the attack accuracy is higher. The reason is that with smaller batch size, poison data tends to be scattered across more training batches resulting in the training difficulty of the adversary, as the mapping between the inputs and features change frequently. Hence the larger batch size in VFL would facilitate privacy attacks.

\subsection{Defense Results}
We compare our proposed defense against the baseline, naive defense, basic defense, and proposed defense variant in terms of the {\em accuracy} and {\em privacy}. Baseline means the VFL without any defense at all, of which the accuracy is typically the highest. In naive defense, we fix the privacy weight factor $\lambda$. In proposed defense variant, we fix stepsizes $\tau_1, \tau_2$ in prior rather than tuning them automatically in proposed defense.

\begin{figure*}[!htbp]
	\centering  
	\subfigure[Purchase, binary classification of $300$ attributes, random guess recall $0.5$ ]{
		\label{fig:defense_purchase}
		\includegraphics[width=0.24\textwidth]{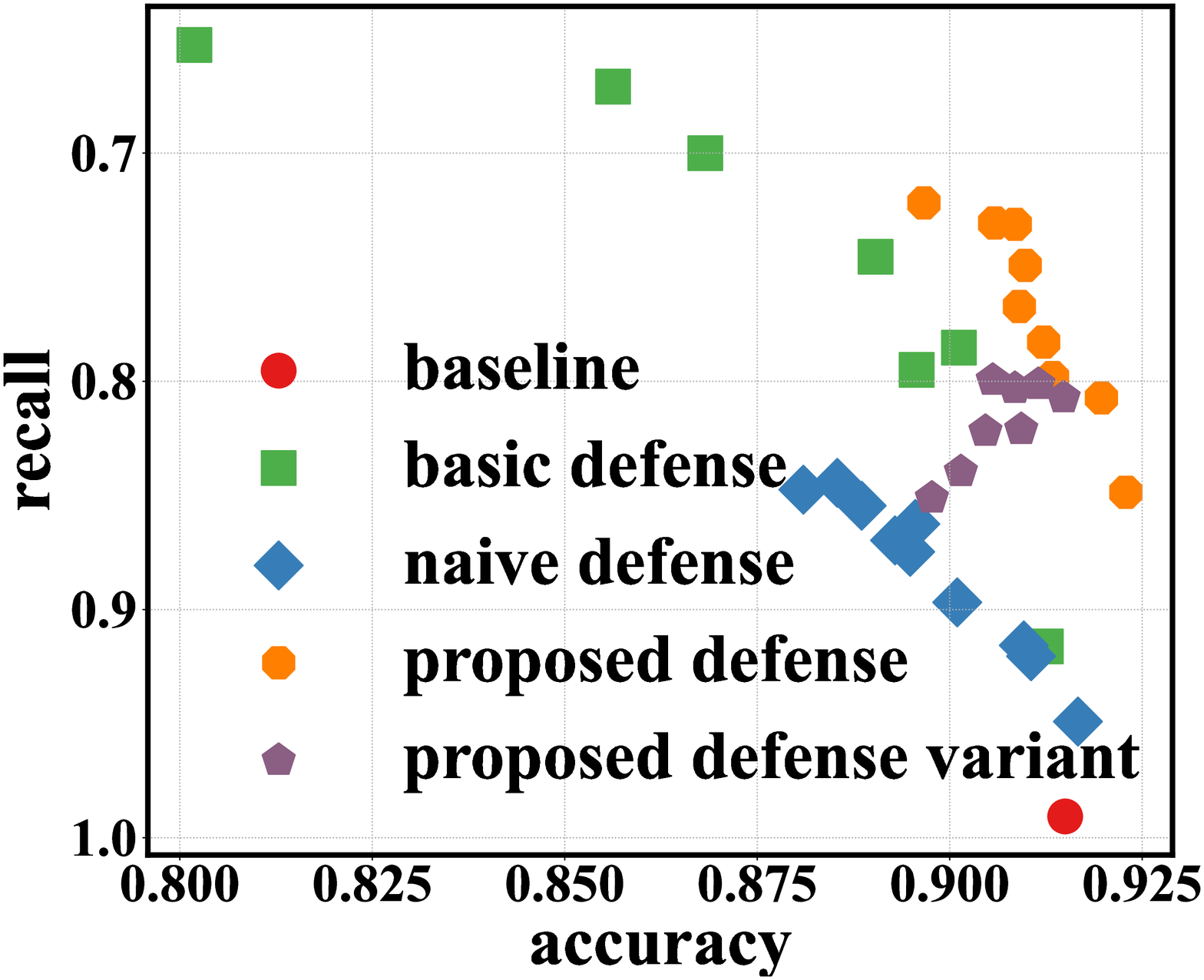}}
	\subfigure[COCO-QA, binary classification on multi-objects, random guess recall $0.0$ due to sparsity]{
		\label{fig:defense_vqa}
		\includegraphics[width=0.24\textwidth]{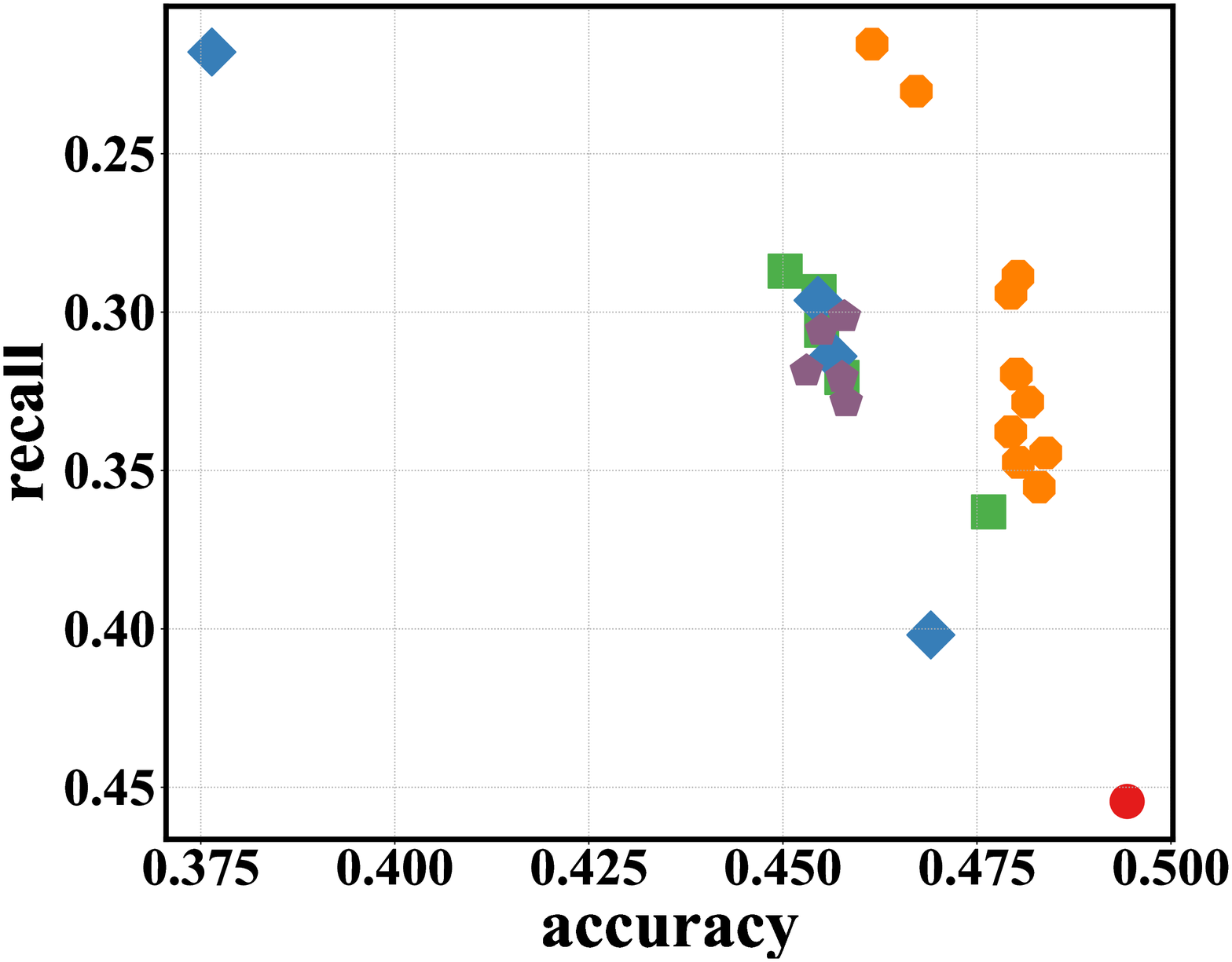}}
	\subfigure[Credit, numerical data reconstruction, random point-wise MSE $2.8 \times 10^{-3}$]{
		\label{fig:defense_credit}
		\includegraphics[width=0.24\textwidth]{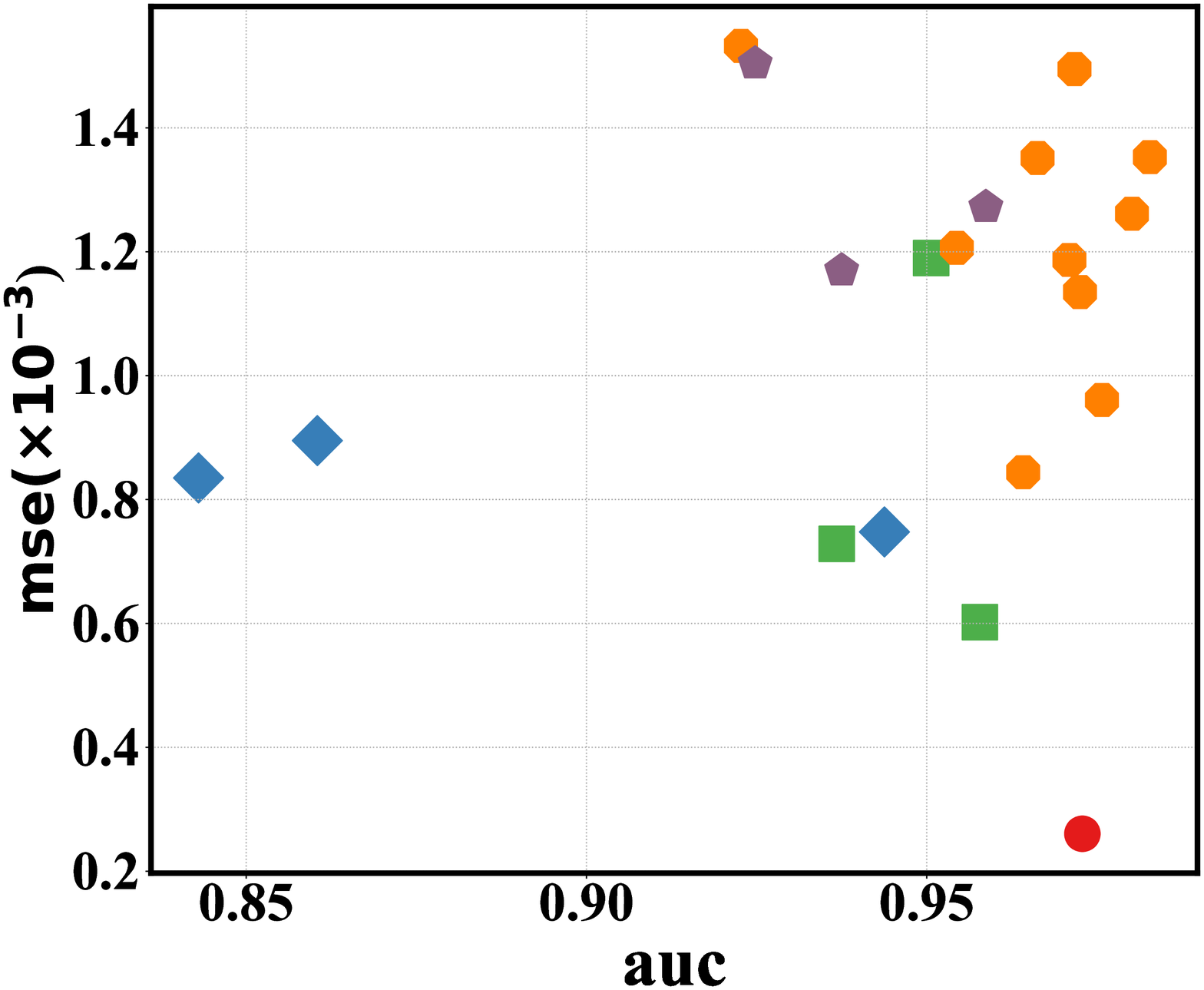}}
	\subfigure[Adult, 1-out-of-15 classification, random guess error rate $14/15$]{
		\label{fig:defense_adult}
		\includegraphics[width=0.24\textwidth]{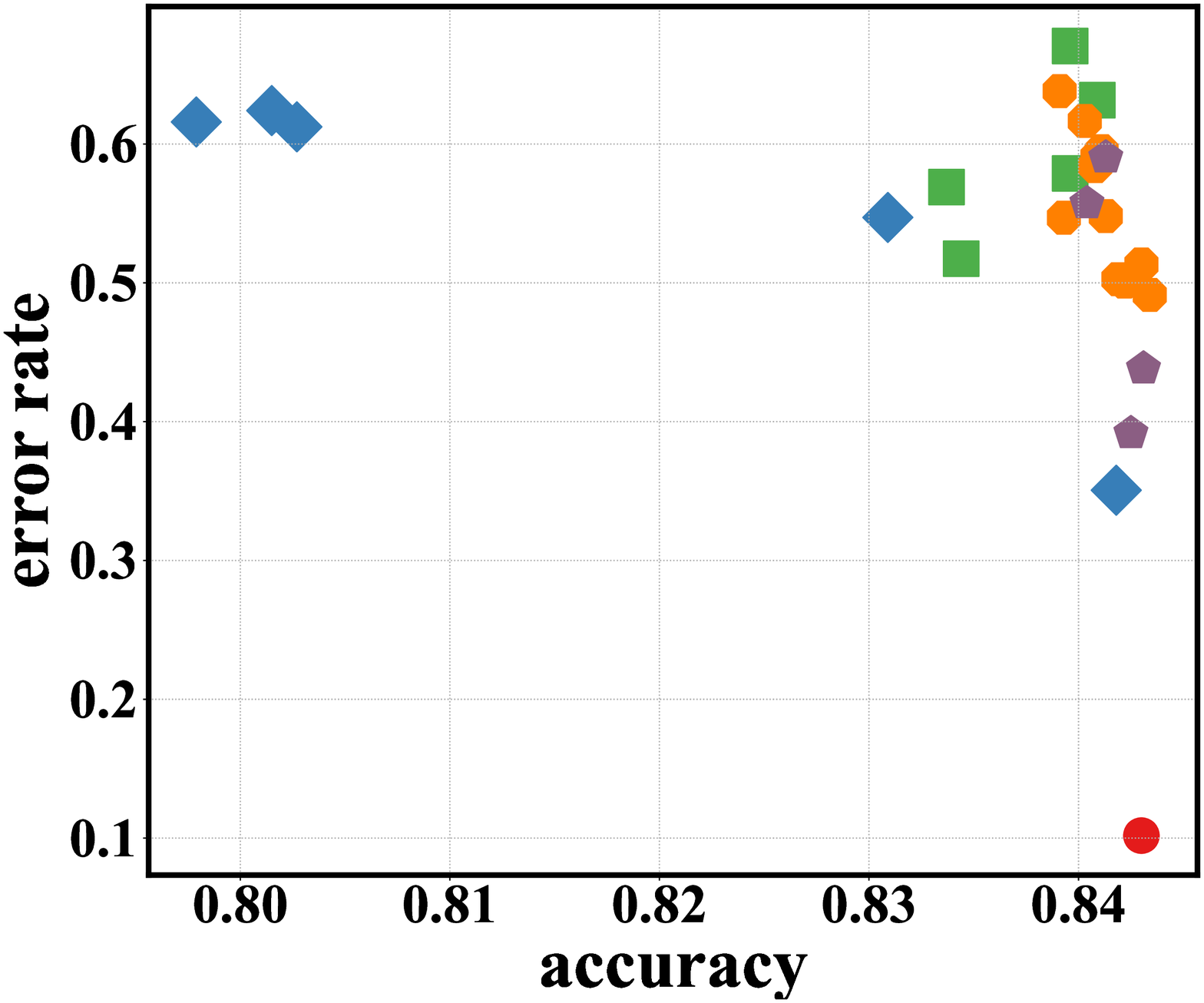}}
	\caption{Defense results on different dataets.  The right and upper corner indicates high accuracy and privacy. All figures share the same legend. We list the dataset name, the specific privacy attack and the success rates of random guesses in each caption. }
	\label{fig:defense}
\end{figure*}

Results on different datasets are shown in Fig.~\ref{fig:defense}. Considering the impact of hyper-parameters, we show multiple results for each method under different settings. Actually, we observe results of the same method are often clustered showing similar performance among all. Baseline achieves the highest accuracy with the lowest privacy. Naive defense and basic defense can achieve high privacy, but at the cost of significant accuracy drop. For example, the naive defense and basic defense respectively achieve the highest privacy with $23.84\%$ and $12.36\%$ accuracy drop in Fig.~\ref{fig:defense_vqa} and Fig.~\ref{fig:defense_purchase}. Our proposed defense is overall superior in that it achieves relatively high privacy with negligible accuracy drop: the average accuracy drop is $0.45\%$ on Purchase, $3.38\%$ on COCO-QA, $0.21\%$ on Adult and $0.69\%$ on Credit. For better understanding of the privacy-preserving effect, we list the success rates of random guess in the caption, which shows the upper limits of privacy. Proposed defense shows significant improvement over baseline, around halfway between baselines and random cases.

{\bf Attribute number.}
We evaluate how the number of private attribute affects the privacy in our proposed defense in Tab.~\ref{tab:purchase_attribute_num}. For example, the client only aims to protect $100, 200, 300$ of its attributes on Purchase. The results are a bit counter-intuitive that as the number of privacy-preserving attributes increases, it is usually harder for the client to prevent privacy leakage. The reason may be the limited capability of local model, not being able to protect all attributes.
\begin{table*}[!htbp]
	\centering
	\scalebox{1.0}{
		\begin{tabular}{ c | c | c | c | c }
			\toprule
			\multirow{2}{*}{Purchase} & Protect attribute number & 100 & 200 & 300 \\ \cline{2-5} 
			& Attack first 100 attributes (Recall)  & 0.785 & 0.7946 & 0.8015 \\ \hline
			\multirow{2}{*}{COCO-QA} & Protect attribute number & 30  & 60  & 90  \\ \cline{2-5} 
			& Attack first 30 attributes (Recall) & 0.489 & 0.4822 & 0.4959 \\ 
			\bottomrule
	\end{tabular}}
	\caption{Protecting different number of attributes: less is better.}
	\label{tab:purchase_attribute_num}
\end{table*}

{\bf Epoch number.}
We also investigate the impact of defense epoch number on accuracy and privacy performance in proposed defense. From Tab.~\ref{tab:defense_epoch_number}, we observe a higher $N_2$ brings lower recall on Purchase and higher error rate on Adult, both indicating higher privacy levels, with little influence on accuracy. It is reasonable that the simulated decoder gets stronger when it is trained for more epochs.
\begin{table*}[!htbp]
	\centering
	\scalebox{1.0}{
		\begin{tabular}{ c | c | c | c | c | c | c }
			\toprule
			Dataset & \multicolumn{3}{c|}{Purchase} & \multicolumn{3}{c}{Adult}  \\ \hline
			$N_2$   & 10  & 20 & 30 & 5 & 10 & 20 \\ \hline
			Accuracy & 0.9134 & 0.9091 & 0.9085 & 0.8394 & 0.8432 & 0.8411 \\ \hline
			Recall / Error rate & 0.7983 & 0.7672 & 0.7312 & 0.2097 & 0.248 & 0.2519  \\ 
			\bottomrule
	\end{tabular}}
	\caption{Defense epoch number $N_2$: more is better.}
	\label{tab:defense_epoch_number}
\end{table*}

\begin{figure*}[!htbp]
	\centering  
	\subfigure[Credit]{
		\label{fig:credit_convergence}
		\includegraphics[width=0.22\textwidth]{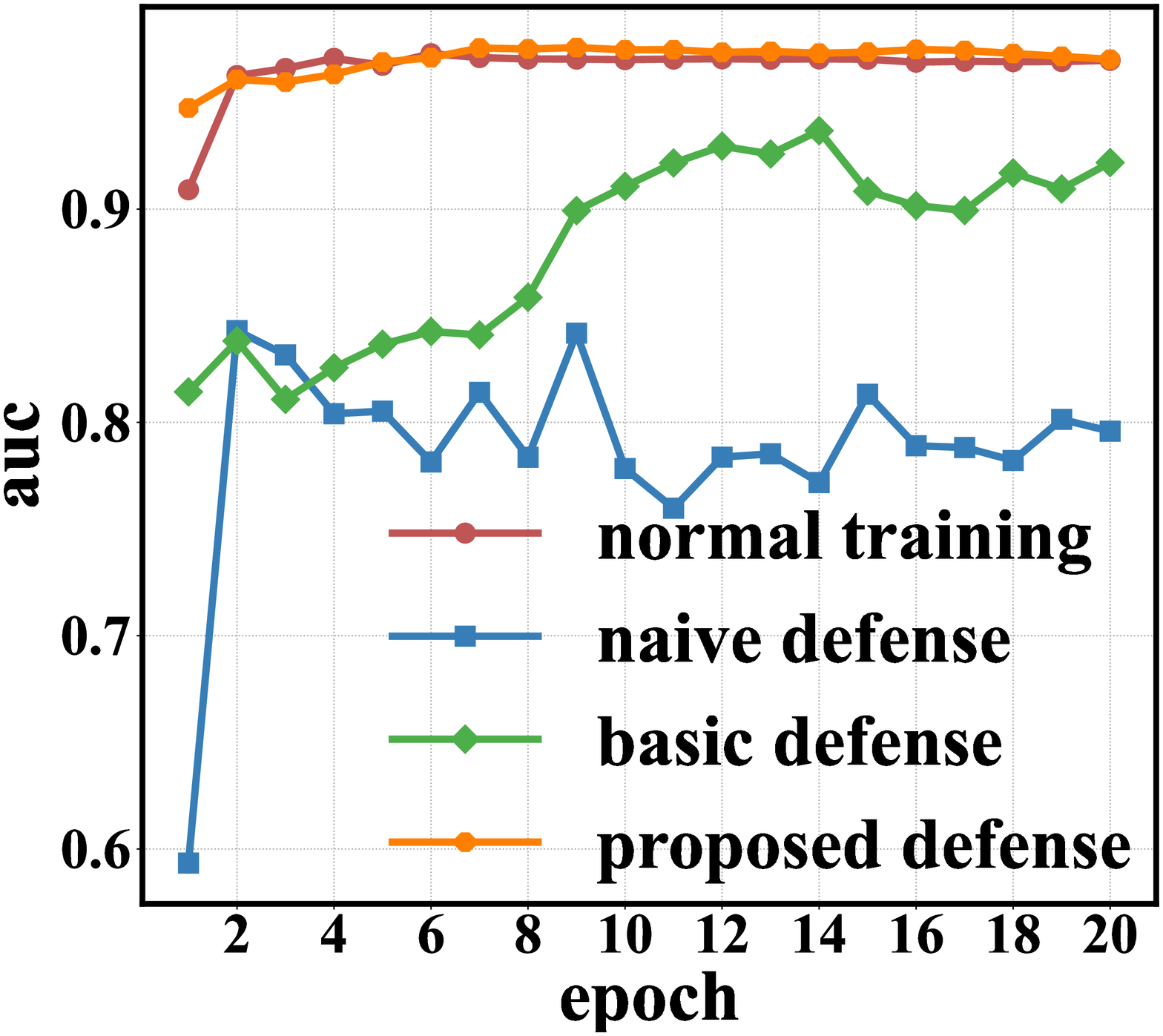}}
	\subfigure[COCO-QA]{
		\label{fig:credit_convergence_}
		\includegraphics[width=0.22\textwidth]{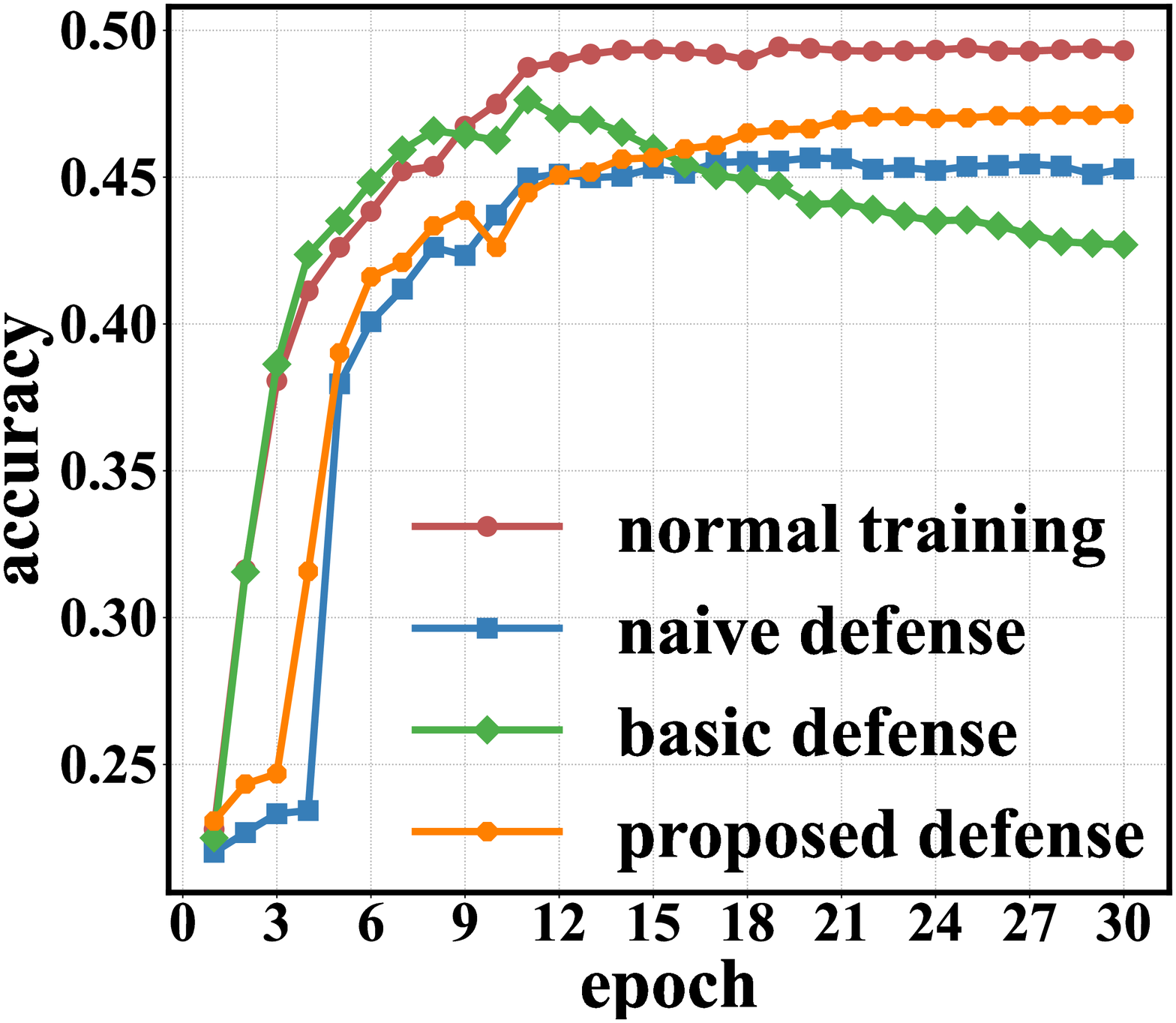}}		
	\caption{Convergence: proposed defense has a similar convergence curve to normal training.}
	\label{fig:convergence}
\end{figure*}

\textbf{Convergence.} Although there are no theoretical guarantees, our proposed defense based on FBS still works well. As shown in Fig.~\ref{fig:convergence}, the convergence curve for proposed defense is much smoother than naive and basic defense with higher accuracy, showing stable training performance, close to the normal one.

\textbf{Multi party.} We show the performance of proposed defense where there are more than 2 clients are involved. Tab.~\ref{tab:mult_purchase} show the case of three clients on Purchase, among which two are victims. Although there is mild accuracy decline compared to 2-client case, the privacy performance is similar, verifying that our approach can be extended to multi-party scenarios.
\begin{table*}[!htbp]
	\centering
	\begin{tabular}{ c | c | c | c | c | c }
		\toprule
		\multirow{2}{*}{$\sim$} & \multirow{2}{*}{Accuracy} & \multicolumn{2}{c|}{Client1, 200 attri} & \multicolumn{2}{c}{Client2, 200 attri} \\ \cline{3-6} 
		&                           & Error rate      & Recall     & Error rate      & Recall     \\ \hline
		Baseline                & 0.9100                    & 0.0654          & 0.9320     & 0.0511          & 0.8925     \\ \hline
		Basic defense           & 0.8790                    & 0.3183          & 0.6760     & 0.3045          & 0.5820     \\ \hline
		Proposed defense        & 0.8879                    & 0.3067          & 0.6876     & 0.3118          & 0.5767     \\
		\bottomrule
	\end{tabular}
	\caption{Three clients on Purchase. $\alpha=0.01.$}
	\label{tab:mult_purchase}
\end{table*}

\section{Conclusion}
VFL enables distributed clients to train model cooperatively over partitioned attributes. We reveal the privacy leakage in the VFL framework, and propose countermeasures based on FBS to split apart the training loops for accuracy and privacy. Our results in a variety of settings show the proposed defense achieves high accuracy and privacy simultaneously with stable training performance.


\bibliographystyle{unsrt}  
\bibliography{main}  

\begin{thebibliography}{10}

\bibitem{konevcny2016federated1}
Jakub Kone{\v{c}}n{\`y}, H~Brendan McMahan, Daniel Ramage, and Peter
  Richt{\'a}rik.
\newblock Federated optimization: Distributed machine learning for on-device
  intelligence.
\newblock {\em arXiv preprint arXiv:1610.02527}, 2016.

\bibitem{konevcny2016federated2}
Jakub Kone{\v{c}}n{\`y}, H~Brendan McMahan, Felix~X Yu, Peter Richt{\'a}rik,
  Ananda~Theertha Suresh, and Dave Bacon.
\newblock Federated learning: Strategies for improving communication
  efficiency.
\newblock {\em arXiv preprint arXiv:1610.05492}, 2016.

\bibitem{mcmahan2017communication}
Brendan McMahan, Eider Moore, Daniel Ramage, Seth Hampson, and Blaise~Aguera
  y~Arcas.
\newblock Communication-efficient learning of deep networks from decentralized
  data.
\newblock In {\em Artificial Intelligence and Statistics}, pages 1273--1282,
  2017.

\bibitem{yang2019federated}
Qiang Yang, Yang Liu, Tianjian Chen, and Yongxin Tong.
\newblock Federated machine learning: Concept and applications.
\newblock {\em ACM Transactions on Intelligent Systems and Technology (TIST)},
  10(2):1--19, 2019.

\bibitem{cheng2019secureboost}
Kewei Cheng, Tao Fan, Yilun Jin, Yang Liu, Tianjian Chen, and Qiang Yang.
\newblock Secureboost: A lossless federated learning framework.
\newblock {\em arXiv preprint arXiv:1901.08755}, 2019.

\bibitem{wu2020privacy}
Yuncheng Wu, Shaofeng Cai, Xiaokui Xiao, Gang Chen, and Beng~Chin Ooi.
\newblock Privacy preserving vertical federated learning for tree-based models.
\newblock {\em arXiv preprint arXiv:2008.06170}, 2020.

\bibitem{zhang2018gelu}
Qiao Zhang, Cong Wang, Hongyi Wu, Chunsheng Xin, and Tran~V Phuong.
\newblock Gelu-net: A globally encrypted, locally unencrypted deep neural
  network for privacy-preserved learning.
\newblock In {\em IJCAI}, pages 3933--3939, 2018.

\bibitem{liu2020federated}
Yang Liu, Yingting Liu, Zhijie Liu, Yuxuan Liang, Chuishi Meng, Junbo Zhang,
  and Yu~Zheng.
\newblock Federated forest.
\newblock {\em IEEE Transactions on Big Data}, 2020.

\bibitem{dosovitskiy2016inverting}
Alexey Dosovitskiy and Thomas Brox.
\newblock Inverting visual representations with convolutional networks.
\newblock In {\em Proceedings of the IEEE conference on computer vision and
  pattern recognition}, pages 4829--4837, 2016.

\bibitem{mahendran2015understanding}
Aravindh Mahendran and Andrea Vedaldi.
\newblock Understanding deep image representations by inverting them.
\newblock In {\em Proceedings of the IEEE conference on computer vision and
  pattern recognition}, pages 5188--5196, 2015.

\bibitem{fredrikson2015model}
Matt Fredrikson, Somesh Jha, and Thomas Ristenpart.
\newblock Model inversion attacks that exploit confidence information and basic
  countermeasures.
\newblock In {\em Proceedings of the 22nd ACM SIGSAC Conference on Computer and
  Communications Security}, pages 1322--1333, 2015.

\bibitem{melis2019exploiting}
Luca Melis, Congzheng Song, Emiliano De~Cristofaro, and Vitaly Shmatikov.
\newblock Exploiting unintended feature leakage in collaborative learning.
\newblock In {\em 2019 IEEE Symposium on Security and Privacy (SP)}, pages
  691--706. IEEE, 2019.

\bibitem{nasr2019comprehensive}
Milad Nasr, Reza Shokri, and Amir Houmansadr.
\newblock Comprehensive privacy analysis of deep learning: Passive and active
  white-box inference attacks against centralized and federated learning.
\newblock In {\em 2019 IEEE Symposium on Security and Privacy (SP)}, pages
  739--753. IEEE, 2019.

\bibitem{goldstein2014field}
Tom Goldstein, Christoph Studer, and Richard Baraniuk.
\newblock A field guide to forward-backward splitting with a fasta
  implementation.
\newblock {\em arXiv preprint arXiv:1411.3406}, 2014.

\bibitem{geiping2020inverting}
Jonas Geiping, Hartmut Bauermeister, Hannah Dr{\"o}ge, and Michael Moeller.
\newblock Inverting gradients--how easy is it to break privacy in federated
  learning?
\newblock {\em arXiv preprint arXiv:2003.14053}, 2020.

\bibitem{zhu2019deep}
Ligeng Zhu, Zhijian Liu, and Song Han.
\newblock Deep leakage from gradients.
\newblock In {\em Advances in Neural Information Processing Systems}, pages
  14774--14784, 2019.

\bibitem{bagdasaryan2020backdoor}
Eugene Bagdasaryan, Andreas Veit, Yiqing Hua, Deborah Estrin, and Vitaly
  Shmatikov.
\newblock How to backdoor federated learning.
\newblock In {\em International Conference on Artificial Intelligence and
  Statistics}, pages 2938--2948, 2020.

\bibitem{luo2020feature}
Xinjian Luo, Yuncheng Wu, Xiaokui Xiao, and Beng~Chin Ooi.
\newblock Feature inference attack on model predictions in vertical federated
  learning.
\newblock {\em arXiv preprint arXiv:2010.10152}, 2020.

\bibitem{geyer2017differentially}
Robin~C Geyer, Tassilo Klein, and Moin Nabi.
\newblock Differentially private federated learning: A client level
  perspective.
\newblock {\em arXiv preprint arXiv:1712.07557}, 2017.

\bibitem{roy2020crypt}
Amrita Roy~Chowdhury, Chenghong Wang, Xi~He, Ashwin Machanavajjhala, and Somesh
  Jha.
\newblock Crypt?: Crypto-assisted differential privacy on untrusted servers.
\newblock In {\em Proceedings of the 2020 ACM SIGMOD International Conference
  on Management of Data}, pages 603--619, 2020.

\bibitem{hardy2017private}
Stephen Hardy, Wilko Henecka, Hamish Ivey-Law, Richard Nock, Giorgio Patrini,
  Guillaume Smith, and Brian Thorne.
\newblock Private federated learning on vertically partitioned data via entity
  resolution and additively homomorphic encryption.
\newblock {\em arXiv preprint arXiv:1711.10677}, 2017.

\bibitem{shokri2017membership}
Reza Shokri, Marco Stronati, Congzheng Song, and Vitaly Shmatikov.
\newblock Membership inference attacks against machine learning models.
\newblock In {\em 2017 IEEE Symposium on Security and Privacy (SP)}, pages
  3--18. IEEE, 2017.

\bibitem{ren2015exploring}
Mengye Ren, Ryan Kiros, and Richard Zemel.
\newblock Exploring models and data for image question answering.
\newblock In {\em Advances in neural information processing systems}, pages
  2953--2961, 2015.

\end{thebibliography}

%
%
%
%

\end{document}